# Using ensemble methods of machine learning to predict real estate prices

Oleh Pastukh [1,†] and Viktor Khomyshyn [1,*,†]

[1] *Ternopil Ivan Puluj National Technical University, Ruska str., 56, Ternopil, 46001, Ukraine*

**Abstract**
In recent years, machine learning (ML) techniques have become a powerful tool for improving the accuracy of predictions and decision-making. Machine learning technologies have begun to penetrate all areas, including the real estate sector. Correct forecasting of real estate value plays an important role in the buyer-seller chain, because it ensures reasonableness of price expectations based on the offers available in the market and helps to avoid financial risks for both parties of the transaction. Accurate forecasting is also important for real estate investors to make an informed decision on a specific property. This study helps to gain a deeper understanding of how effective and accurate ensemble machine learning methods are in predicting real estate values. The results obtained in the work are quite accurate, as can be seen from the coefficient of determination ($R^2$), root mean square error (RMSE) and mean absolute error (MAE) calculated for each model. The Gradient Boosting Regressor model provides the highest accuracy, the Extra Trees Regressor, Hist Gradient Boosting Regressor and Random Forest Regressor models give good results. In general, ensemble machine learning techniques can be effectively used to solve real estate valuation. This work forms ideas for future research, which consist in the preliminary processing of the data set by searching and extracting anomalous values, as well as the practical implementation of the obtained results.

## 1. Introduction

One of the sectors of the economy, which is the most attractive for investments by individuals and businesses, is always real estate [1]. In this market, buyers are always looking to get the best option for the lowest possible price, while the seller is looking to get the most out of their property. At the same time, the starting price of the object in the vast majority is overestimated due to the inability of the seller to conduct a comprehensive analysis of prices for similar objects, many objects offers and their characteristics. Many real estate agents and rector companies in this case offer the seller a price based on their practical experience. This version of real estate valuation is far from optimal. In the process of searching for housing, it is extremely important for a buyer to accurately analyze his personal budget and available prices in relation to the desired characteristics of real estate. There is a need to develop models and tools that provide an accurate analysis of existing market prices with the possibility of evaluating new real estate properties according to their characteristics [2]. The field of real estate services is interested in appropriate convenient services and software that provide the necessary functionality for making informed decisions regarding the sale and purchase of real estate [3]. Machine learning methods, which have been actively developing in recent years, come to the rescue [4].

## 2. Related works

The field of forecasting is a part of machine learning technologies, which allows, through the application of various models, optimized for certain tasks, to obtain approximate predicted price values with the help of machine learning. The use of machine learning methods for forecasting sales prices is considered in the works of scientists from many countries. Research [5] provides valuable insight into the complex dynamics of property prices and the effectiveness of predictive modeling in understanding and predicting these trends in the UK property market, highlighting the importance of adopting a multi-dimensional approach to property valuation that includes both quantitative indicators and qualitative findings. In [6], by applying machine learning models, the largest real estate data set in Germany, which contains more than 1.5 million unique samples and more than 20 characteristics, is investigated. The obtained results indicate that the forecast of sales prices in a real scenario is achievable but limited by the quality of the data. A comparison of machine learning regression methods that produce successful forecasting results using a June-July 2021 property sales dataset in Ankara Province, Turkey indicates that XGBoost and Artificial Neural Networks (ANN) methods are valuable alternatives for predicting property sales prices [7].

Several studies in predicting housing prices based on machine learning algorithms use, in particular, Random Fores and Linear Regression [8], Random Fores and XGBoos [9], Random Fores, Extreme Gradient Boosting, Light Gradient Boosting Machine and Hybrid Regression [10]. In [11], a Django-based REALM web application for automating the real estate valuation process using the Random Fores machine learning model is presented. This model was chosen as the default for the platform over the Neural Network and Linear Regression models due to the higher R2 value obtained when testing the data set and because of the relative simplicity of implementation. Optimization techniques analyzed included tuning hyperparameters and adding a batching layer to the default model. With the help of optimization methods, the model achieved an R2 score of up to 0.9 and higher. Comparative analysis of machine learning algorithms for forecasting real estate prices is published in works [12], [13], [14], [15]. The cloud-based application for forecasting housing prices using machine learning proposed in [16] uses the XGBoost regressor, which achieved the best results among the five studied machine learning algorithms - linear regression (LR), decision tree (DT), random forest (RF), additional tree regressor (ETR) and extreme gradient regressor (XGBoost).

The successful results obtained by various actors in the presented studies encourage the expansion of the used machine learning methods and the use of complex models based on existing simple forecasting algorithms. This is achieved by creating ensembles of models combining several learning algorithms. Ensembles provide more accurate results and improve model stability and performance. They can be created in different ways: bagging, boosting, voting and stacking. In this case, several weak algorithms are combined according to a special pattern to create a stronger one that works better than several weaker ones individually.

The number of scientific works covering ensemble methods of machine learning in forecasting real estate prices is small. The work [17] proposes a combination of machine learning algorithms and statistical methods. In the proposed model, the k-means clustering algorithm clusters the data. Random Forest extracts prediction trees. Shrinkage techniques such as lasso, elastic net, and group lasso are used to reduce the number of trees in a random forest. The generated trees are fed into lasso, group lasso, and elastic mesh algorithms for tree reduction. Finally, the remaining trees are assembled into an ensemble.

The Ada Boost Regressor, Bagging Regressor, Stacking Regressor, and Voting Regressor ensemble models were analyzed using a publicly available dataset downloaded from the Kaggle website [18]. The dataset contains up-to-date information on homes of various properties in the city of California. According to the obtained results, the AdaBoost Regressor model received the best performance value of 0.118 RMSE. The study also created a website to estimate the value of a user's home. [19] analyzed house price data obtained from a leading Thai real estate website and Open Street Maps (OSM). To determine the most efficient stacking model for this data set, the authors evaluated the performance of the RF, XGBoost, and AdaBoost models, which are powerful ensemble-based models. The experimental results showed that the AdaBoost model can provide better performance in terms of MAPE. In addition, the CNN-XGBoost stacking model was found to outperform other individual methods. Research [20] notes that after applying machine learning methods (linear regression, KNN, SVM) and ensemble methods (XGB regressor, AdaBoost, gradient boost regressor, random forest, CatBoost), the latter give better results and accuracy, compared to traditional models, for the Ames and King County public datasets. In ensemble learning techniques, the CatBoost method takes first place. When determining the accuracy of each of the models, the RMSE, MSE, MAE and $R^2$ metrics were considered.

## 3. Research methodology

### 3.1. Data collection

To obtain and maintain the relevance of the data set, the authors developed software in the form of a Windows application and a Microsoft SQL Server database [21]. This software allows you to automate the work of real estate agencies and includes functionality that, by parsing Internet pages, collects data on real estate objects from the DIM.RIA [22] and OLX.ua [23] sites. In general, we collected data published for the period from January 2022 to July 2024 for analysis. The data set contains a description of 1200 offers for the sale of apartments and rooms in the city of Ternopil, Ukraine. The relevance of each object, as well as the characteristics missing from the ad, were clarified with the owner. Objects advertised by real estate agencies were not added to the database, because commission agencies may be included in the price of such objects, which can distort the accuracy of the price forecast. To form a data set that will be used for machine learning in the future, the software was supplemented with the function of exporting the necessary fields to the CSV format.

### 3.2. Dataset Description

Modern real estate portals suggest entering up to 200 of its descriptive characteristics when describing objects, of which up to 15 are mandatory [24]. The software developed by us for the formation of a database contains about 50 parameters when describing apartments and rooms. In both cases, only a part of these parameters has a decisive influence on the price. The data set that we export from the machine learning analysis program contains 12 columns. The first column is the ID of the object, which we exclude from further processing, since it serves only as a unique identifier and has no intrinsic predictive value. The next 10 columns are characteristics that describe the property. The last column is the price of the object, which is the target variable (function) for prediction. A detailed description of the data set is given in table 1.

**Table 1**
Dataset description

| Feature ID | Feature name | Feature type | Description |
|---|---|---|---|
| 0 | id | integer | The ID of the object in the database |
| 1 | realty_type | text | Type of real estate |
| 2 | total_area | real | Total area ($m^2$) |
| 3 | floor | integer | The floor on which the real estate is located |
| 4 | floors | integer | Total number of floors in the building |
| 5 | repair_state | text | State of repair |
| 6 | wall_material | text | Material of external walls |
| 7 | furniture | text | Availability of furniture |
| 8 | heating | text | Type of heating |
| 9 | build_year | text | Construction years |
| 10 | market | text | Real estate market |
| 11 | price | integer | The price of the object ($) |

### 3.3. Research procedure

Data analysis was performed in the Python programming language using the Spyder IDE tool included in the Anaconda software package. This platform specializes in scientific computing, in particular, in the application of machine learning, analytics and data processing methods. Data processing was performed using the pandas, numpy and sklearn libraries, visualization - using the matplotlib library. After uploading information from the

csv file to the Python program, the data set was pre-processed, which included the following steps:

- removal of non-informative "id" column
- removing duplicate lines
- deleting columns with missing data
- coding of categorical features (text values)
- dividing the dataset into an array of independent 'x' and dependent 'y' variables
- dividing the dataset into training (75%) and test (25%) sample

Next, research and analysis of ensembles of machine learning methods, which currently contain one of the most popular machine learning libraries Scikit-learn [25] with basic configuration parameters, was carried out, and the performance of each model was evaluated based on the metric – coefficient of determination ($R^2$), root mean square error (RMSE) and mean absolute error (MAE).

## 4. Results

The actual and predicted values of real estate prices, constructed using the considered ensemble methods of machine learning, are presented in Fig. 1. In the table 2 shows the evaluation metrics of each of the models. A comparison of the values of each of the metrics is presented in Fig. 2-4. The obtained results demonstrate that the Gradient Boosting Regressor model performs best among all models with $R^2$=0.724, RMSE=11,980, MAE=8,113. The Extra Trees Regressor, Hist Gradient Boosting Regressor, and Random Forest Regressor models provide close and also good results at the level of $R^2$=0.688...0.696, RMSE=12,563...12,745, MAE=8,691...8,836. As for the rest of the models, their forecast results are satisfactory and require further analysis.

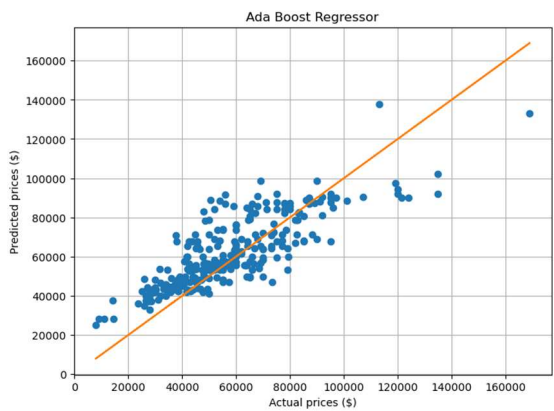
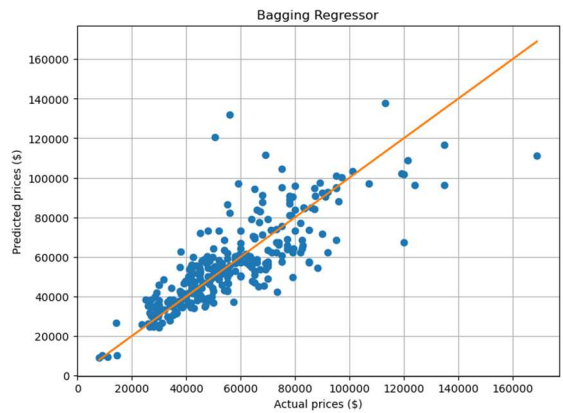

a) Ada Boost Regressor        b) Bagging Regressor

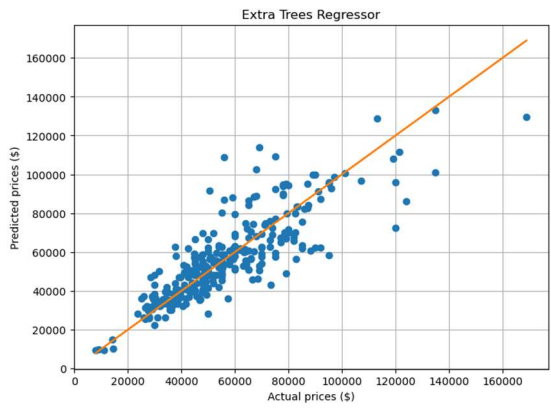
c) Extra Trees Regressor

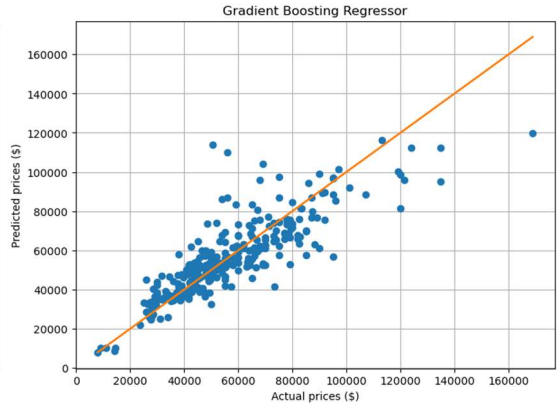
d) Gradient Boosting Regressor

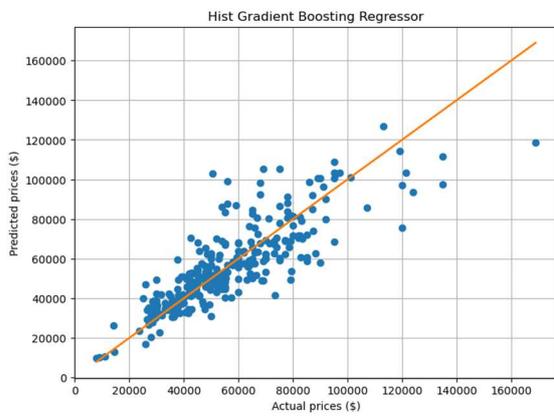
e) Hist Gradient Boosting Regressor

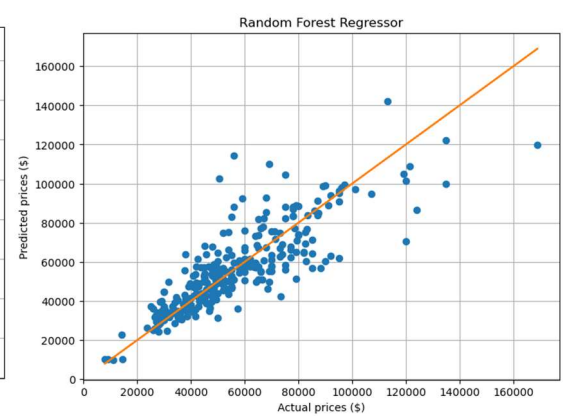
f) Random Forest Regressor

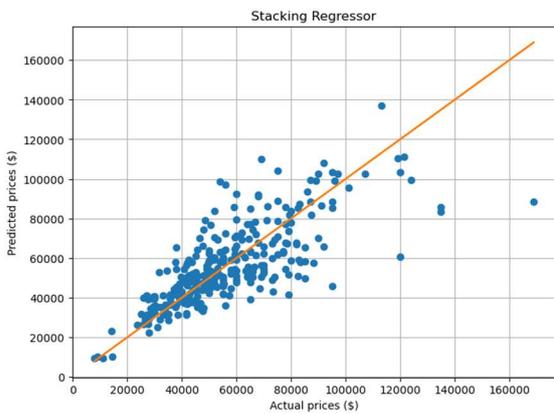
g) Stacking Regressor

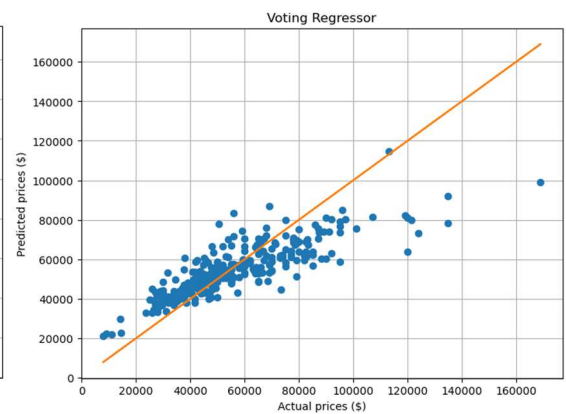
h) Voting Regressor

**Figure 1:** The actual and predicted values prices of real estate, built by means ensemble methods machine learning

**Table 2**
The evaluation metrics of models

| Model | R² | RMSE | MAE |
|---|---|---|---|
| Ada Boost Regressor | 0.641 | 13 664 | 10 739 |
| Bagging Regressor | 0.636 | 13 763 | 9 174 |
| Extra Trees Regressor | 0.696 | 12 563 | 8 691 |
| Gradient Boosting Regressor | 0.724 | 11 980 | 8 113 |
| Hist Gradient Boosting Regressor | 0.696 | 12 576 | 8 836 |
| Random Forest Regressor | 0.688 | 12 745 | 8 731 |
| Stacking Regressor | 0.571 | 14 932 | 10 612 |
| Voting Regressor | 0.635 | 13 780 | 10 066 |

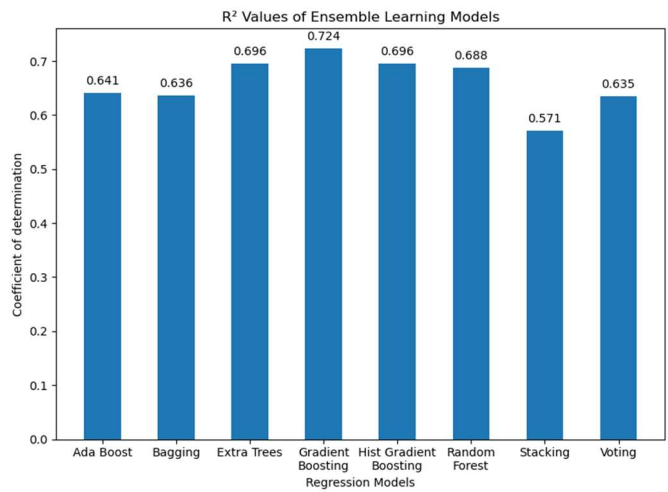

**Figure 2:** R² Values of Ensemble Learning Models

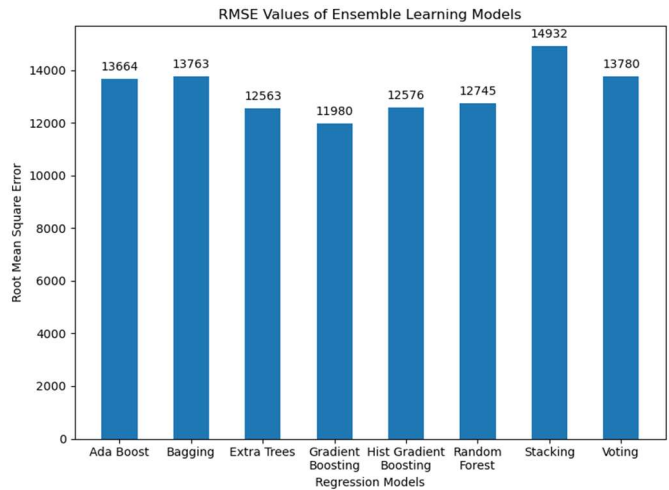

**Figure 3:** RMSE Values of Ensemble Learning Models

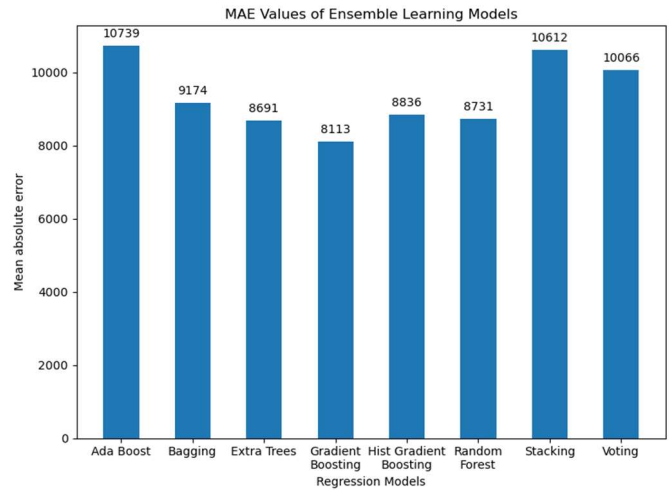

**Figure 4:** MAE Values of Ensemble Learning Models

## 5. Conclusions

Our research has shown that ensemble machine learning algorithms show good potential in real estate valuation. Among the eight considered algorithms, the Gradient Boosting Regressor provides the highest accuracy, the Extra Trees Regressor, Hist Gradient Boosting Regressor and Random Forest Regressor algorithms give a good result. The real estate price forecast itself is still not sufficiently accurate, it can be interpreted as a preliminary one with a tendency to gradually improve the methodology. The reasons for this lie partly in the data itself than in the applied machine learning methods. Correlation of characteristics with sales price shows that not all factors that determine price formation are present in the data. The task and goal of our future research is to improve the accuracy of the assessment of ensemble methods by analyzing datasets for anomalous values, adjusting the hyperparameters for the studied ensemble algorithms that showed the best results, supplementing the dataset with more detailed information about real estate, including geospatial coordinates of real estate objects, practical implementation and testing of the researched models. Ultimately, our work aims to develop a multi-agent real estate evaluation system based on modern innovative machine learning algorithms.